\documentclass[runningheads]{llncs}
\usepackage{graphicx}
\usepackage{amsmath}
\usepackage{booktabs} 
\usepackage{amssymb}
\usepackage{wrapfig}
\begin{document}
\title{Adaptive Discrete Disparity Volume for Self-supervised Monocular Depth Estimation}
\titlerunning{ADDV for Self-supervised Monocular Depth Estimation}
\author{Jianwei Ren}
\authorrunning{}
\institute{
\email{jvren42@gmail.com}}
\maketitle              % typeset the header of the contribution
\begin{abstract}
In self-supervised monocular depth estimation tasks, discrete disparity prediction has been proven to attain higher quality depth maps than common continuous methods. However, current discretization strategies often divide depth ranges of scenes into bins in a handcrafted and rigid manner, limiting model performance. In this paper, we propose a learnable module, Adaptive Discrete Disparity Volume (ADDV), which is capable of dynamically sensing depth distributions in different RGB images and generating adaptive bins for them. Without any extra supervision, this module can be integrated into existing CNN architectures, allowing networks to produce representative values for bins and a probability volume over them. Furthermore, we introduce novel training strategies - uniformizing and sharpening - through a loss term and temperature parameter, respectively, to provide regularizations under self-supervised conditions, preventing model degradation or collapse. Empirical results demonstrate that ADDV effectively processes global information, generating appropriate bins for various scenes and producing higher quality depth maps compared to handcrafted methods.
\end{abstract}
\section{Introduction}
Human beings are competent in estimating the relative depth of a scene even conditional on only a single image. This vision-based capacity is also of vital crucial for autonomous driving and robotic systems for its prediction of dense depth maps. Recent years, in order to prevent from using expensive LiDAR data or other complex sensors, learning-based self-supervised methods have been well developed and showing its prosperity.

Predicting depth with a discrete method in monocular depth estimation (MDE), which recasts it as a classification problem, has been proven by numerous studies \cite{bhat2021adabins,cao2017estimating,fu2018deep,johnston2020self,liu2019neural} to be superior to a common continuous one. Considering MDE as a classification task, rather than a regression one, helps generate sharp and high-quality depth maps \cite{johnston2020self}. It also facilitates measuring depth uncertainty in scenes \cite{liu2019neural}, which is of great significance for system security. In this paper, we therefore adopt this notion that dividing the depth range into bins and predicting the probability distribution over them instead of directly regressing a single depth value for each individual pixel. 

There are two handcrafted strategies existing: uniform discretization (UD) and spacing-increasing discretization (SID). The representative values of UD, which are referred as to the center value of bins, grows linearly, meaning that UD splits the depth range into uniform intervals. SID, on the other side, is to uniformly quantize in $log$ space, indicating the exponential growth of widths. According to \cite{fu2018deep}, SID works better than UD because an image usually bears less information for larger depth. 

However, as the depth distribution varies dramatically from scene to scene, both UD and SID's rigidity impair model's performance because they treat distinct inputs indiscriminately. Recent attempts by \cite{bhat2021adabins,li2022binsformer} to solve this problem still require explicit supervision. In this paper, we wonder if the adaptive strategy could remain effective and facilitate the processing of global information in the absence of ground truth.

We propose a learnable module called Adaptive Discrete Disparity Volume (ADDV), which enables a network to dynamically generate bins and estimate probability distributions for samples according to input images. It has been observed that the absence of supervised training results in instability: the model lacks guidance on adjusting the width of bins to adapt to different scenes. To address this issue, we propose utilizing sample balancing as the criterion for adaptive bin generation, referred to as uniformizing. It compels the network to adjust bins to ensure even distribution of samples within them. Additionally, we stimulate extreme values in the probability distribution of individual samples over bins, referred to as sharpening, which mitigates bias introduced by multimodal distributions while maintaining differentiability. 

Finally, experimental results demonstrate that the model with ADDV outperforms UD and SID under self-supervised conditions, yielding higher-quality depth maps. Our ablation studies prove the effectiveness of the two aforementioned strategies to the performance of the adaptive method.

\begin{figure}
    \centering
    \includegraphics[width=\textwidth]{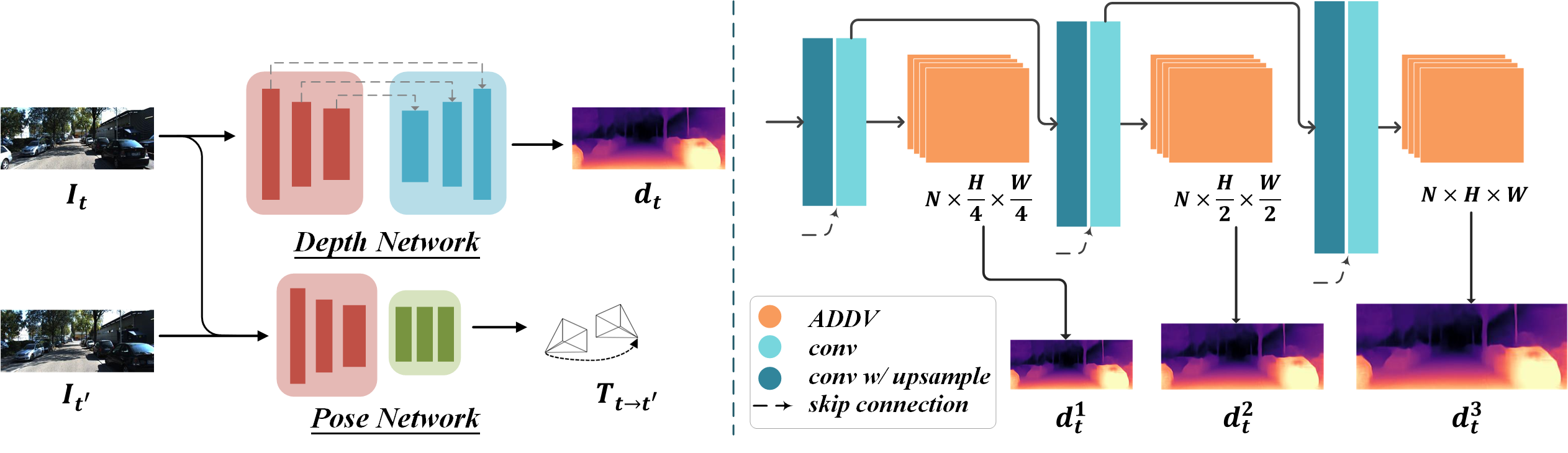}
    \begin{tabular}{c@{\hskip 60mm}c}
    \small{a)} & \small{b)} \\
    \end{tabular}
    \caption{\textbf{Overview.} a) The framework consists of an encoder-decoder depth estimation network and a separate pose estimation network. b) Detailed decoder of the depth net. ADDV modules are inserted to achieve adaptive depth discretization.}
\label{fig1}
\end{figure}

\section{Related Work}
\subsection{Self-supervised Monocular Depth Estimation}
Recovering scene depth from a single image is a challenging problem in computer vision, particularly in self-supervised settings where depth information is lost during image capture, making it inherently ill-posed. In recent years, many learning-based methods develop their models via epipolar geometry constraints \cite{godard2017unsupervised}, reconstructing a target frame by warping a source frame inversely to guide training via reconstruction loss \cite{garg2016unsupervised,zhou2017unsupervised}. Zhou et al. \cite{zhou2017unsupervised} present a purely unsupervised framework trained on consecutive temporal frames, jointly estimating depth and 6-DoF pose from ego-motion. The notion of view synthesis is taken further by \cite{godard2019digging}, which proposes an advanced multi-scale framework with a novel minimum reprojection loss and auto-masking. Both improvements assist in coping with the exceptional regions of images where the assumption of photometric consistency is violated, such as occlusions and non-Lambertian surfaces.

Some works introduce extra clues to provide more priors. For example, \cite{ranjan2019competitive,yin2018geonet,zou2018df} propose networks with joint learning of depth and optical flow, while \cite{bian2019unsupervised,mahjourian2018unsupervised,yang2018lego,yang2018unsupervised} leverage geometry constraints. Semantic information could be a more popular prior, which has been utilized either to model motion \cite{casser2019depth,gordon2019depth,lee2021attentive} or for multi-task models \cite{chen2019towards,choi2020safenet,guizilini2020semantically,klingner2020self}. However, these approaches also increase model complexity and hence cause higher computational costs. In contrast, our method requires no additional clues with lower expenses.  

\subsection{Depth discretization}
During imaging, a camera's 3D view frustum can be hierarchically divided so that each point in the scene will fall in an interval or bin with a representative depth value. This discretization process has inspired numerous studies \cite{bhat2021adabins,cao2017estimating,fu2018deep,johnston2020self,liu2019neural} to assign image pixels into bins with associated probabilities, instead of directly regressing specific depth values. These pixel probability distributions across bins form a volume, leading to the final depth map being computed as a linear combination \cite{bhat2021adabins,cao2017estimating,johnston2020self,liu2019neural} of representative values weighted by the probability volume.

Two prevalent discretization strategies have emerged: uniform discretization (UD) and spacing-increasing discretization (SID). SID was initially introduced in \cite{cao2017estimating} and was found by \cite{fu2018deep} to outperform UD in accuracy. However, as noted by \cite{bhat2021adabins}, the depth distribution varies significantly across different images, posing a challenge for both SID and UD to adapt simultaneously. To address this issue, \cite{bhat2021adabins} proposes adaptive bins, whose widths can change according to inputs, albeit in a supervised manner. In the context of self-supervised learning, fewer studies exist, but \cite{johnston2020self} introduces discrete disparity volume using conventional discretization strategies.

In this paper, we propose a purely self-supervised adaptive strategy. We argue that both UD and SID represent specific instances of adaptive strategies.

\section{Method}
\subsection{Problem Formulation}
To obtain supervision signals from temporally continuous view transformations, our framework jointly optimizes both a depth prediction network and a pose one. The depth net takes an RGB frame $I_t\in \bbbr^{3\times H\times W}$ as input and produces its corresponding depth map $d_t\in \bbbr^{H\times W}$, denoted as $\mathcal{D}(I_t)\rightarrow d_t$. At the same time, the pose network receives a concatenation of two adjacent frames $I_t$ and $I_{t'}\in \{I_{t-1},I_{t+1}\}$ and outputs 6-DoF relative pose $T_{t\rightarrow t'}\in SE(3)$, denoted as $\mathcal{P}(I_t, I_{t'})\rightarrow T_{t\rightarrow t'}$. Through the warping operation, expressed as $\varphi (I_{t'}, d_t, T_{t\rightarrow t'}, K)\rightarrow I_{t'\rightarrow t}$, where $K$ represents the camera intrinsics, a reconstructed frame $I_{t'\rightarrow t}$ is obtained. Similar to \cite{godard2019digging}, the primary objective is to optimize a photometric error $pe$ of reconstruction: 
\begin{equation}
    pe(I_1, I_2) = \frac{\alpha _1}{2}(1 - SSIM(I_1, I_2)) + (1 - \alpha _1)||I_1 - I_2||
\end{equation}
where $\alpha_1$ is a scale factor. With assistance of auto-masking $\mathcal{M}(\cdot)$, a pixel-wise minimum reprojection loss is employed to lessen the effects of occlusions, static pixels and low-texture regions: 
\begin{equation}
     L_p=\mathcal{M}(\mathop{min}_{t'} pe(I_t, I_{t'\rightarrow t}))
\end{equation}
Additionally, an edge-aware smoothness term $L_s$ is used to reduce artifacts:
\begin{equation}
    L_{smooth} = |\partial_xd_t^*|e^{-|\partial_xI_t|} + |\partial_yd_t^*|e^{-|\partial_yI_t|}.
\end{equation}
where $d_t^*=d_t/\bar{d_t}$ is the mean-normalized disparity to improve numerical stability \cite{wang2018learning}.

Figure 1 illustrates the framework and a detailed decoder of the depth network. The decoder involves five blocks, with the ADDV incorporated into four of them to generate depth maps for each scale, excluding the lowest-resolution block.

\begin{figure}
    \includegraphics[width=\textwidth]{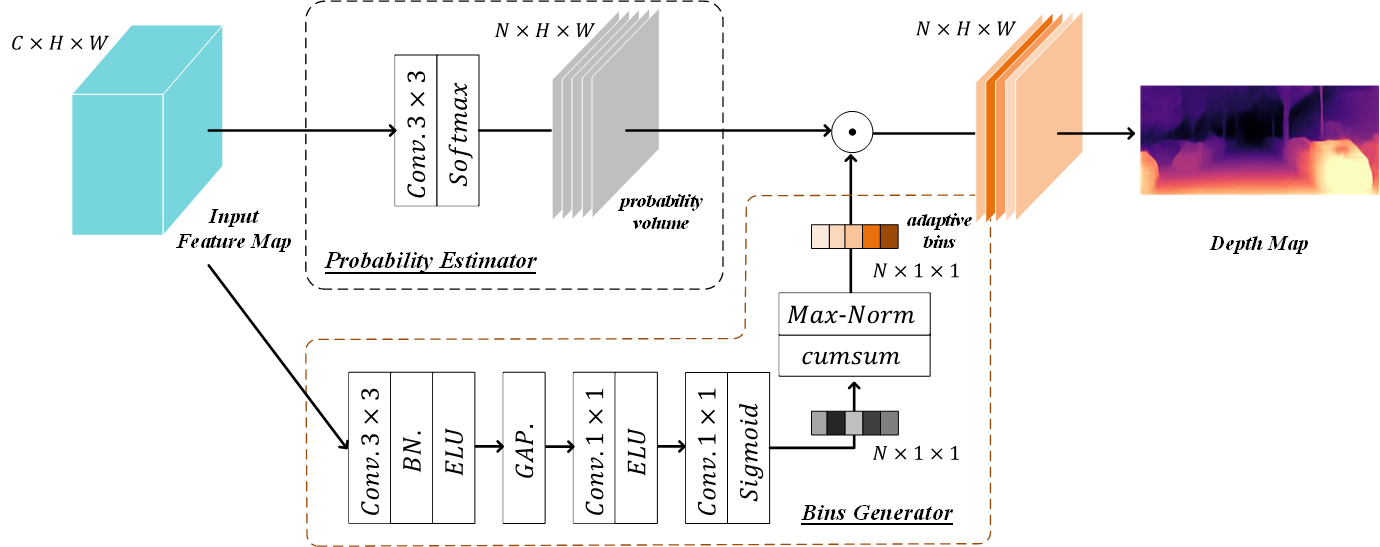} 
    \caption{\textbf{Detailed ADDV.} The upper component predicts probability distributions of pixels and aggregates them into a volume, while the lower is designed to generate adaptive bins.}
\end{figure}

\subsection{Adaptive Discrete Disparity Volume}
Adaptive discrete disparity volume (ADDV) is a differentiable module that is included in each output block of the depth decoder, applicable at any scale due to its resolution insensitivity. This module predicts a probability volume and generates adaptive bins conditional on the high-dimension features from the last layer. Subsequently, the bin values, representing relative depth, are weighted by probabilities and aggregated to produce the final depth map.

ADDV consists of two components, as depicted in Figure 2. A \textit{probability estimator} predicts and aggregates per-pixel probability distributions, forming a volume from an input feature map $X\in \bbbr^{C\times H\times W}$  with $C$ channels and resolution $H\times W$. To meet the quantization requirement of discrete methods, the input $X$ is processed with convolutional kernels, resulting in a logit feature map $Y\in \bbbr^{N\times H\times W}$.  The $softmax$ is applied to yield well-defined per-pixel probabilities: 
\begin{equation}
    P^n_{(u,v)}=softmax(Y)\equiv \frac{\exp (y^n_{(u,v)})}{\sum_{i=1}^N\exp (y^i_{(u,v)})}
\end{equation}
where $y^n_{(u,v)}$ is the value of pixel $(u,v)$ in the $n$-th channel of $Y$.

As the other component of ADDV, the \textit{bins generator} captures depth clues and produce adaptive bins accordingly. It initially convolves the input feature map to align with the probability volume across channels. To incorporate global context, global average pooling integrates spatial information, resulting in an $N\times 1\times 1$ tensor that encapsulates implicit information about bin widths. The last convolutional layer is activated by $sigmoid$, ensuring that relative widths remain positive within a reasonable range. Finally, a cumulative sum operator and a max normalization are imposed on the tensor to obtain the representative values of adaptive bins, denoted as $\textbf{b}=[b^1,b^2,...,b^N]$, where $b^n$ refers to the $n$-th element. 

A common approach is to compute the final depth map $d\in \bbbr^{H\times W}$ from probability volume using Maximum Likelihood Estimates (MLE), formally as: 
\begin{equation}
    d_{(u,v)}=b^{n^*};n^*=\mathop{argmax}_nP^n_{(u,v)}
\end{equation}
where $d_{(u,v)}$ stands for the depth value of pixel $(u,v)$. However, it fails to converge due to ${argmax}$'s non-differentiability. A modified MLE, adopted by \cite{fu2018deep}, overcomes this limitation but still introduces artifacts in depth maps \cite{bhat2021adabins}. Consequently, most discrete methods \cite{bhat2021adabins,johnston2020self,liu2019neural} resort to \textit{soft-argmax}, in a form of linear combinations of probability volume and representative values: 
\begin{equation}
    d_{(u,v)}=\sum_{n=1}^Nb^n\cdot P^n_{(u,v)}
\end{equation}
In ADDV, we implement this by performing an element-wise multiplication between the broadcast adaptive bins and the probability volume.

\subsection{Uniformizing and Sharpening}
Models employing adaptive strategies excel at global information processing. However, generating adaptive bins poses an extra task and burden during training. Previous works \cite{bhat2021adabins,li2022binsformer} rely on supervision from ground truth to prevent performance degradation. In this paper, to achieve this in a self-supervised environment, we propose leveraging  the principle of sample balancing as a prior for adaptive bin adjustment. Moreover, to mitigate the estimation bias caused by the \textit{soft-argmax} operation, we employ sharpening to stimulate peaks in the probability distributions.

\textbf{Uniformizing} loss term promotes adjustments to bin widths to ensure a balanced distribution of samples across bins. We provide two designs to implement this term. An intuitive but naive one $L_u^{i)}$ is defined as: 
\begin{equation}
\begin{aligned}
    &c_n=\sum_{u,v}\eta (\mathop{argmax}_iP^i_{(u,v)}=n)\\
    &L_u^{i)}=\sum^N_{n=1}\|\frac{c_n}{c_{valid}}-\frac{1}{N}\|
\end{aligned}
\end{equation}
where $\|\cdot\|$ denotes the $L2$ norm and $c_{valid}$ is the number of valid samples. The indicator function $\eta(\cdot)$ equals 1 if the inner condition is true, and 0 otherwise. In this formulation, $c_n$ represents the number of samples determined by MLE in the $n$-th bin. However, this term potentially hinders training as it only considers the indices of max value in each probability distribution and discards much of the available information. 

An improved alternative $L_u^{ii)}$ is proposed to take into account the entire probability distributions of all samples. It can be written as: 
\begin{equation}
\begin{aligned}
    &P^n_{avg}=\frac{1}{c_{valid}}\sum_{u,v}P^n_{(u,v)}\\
    &L_u^{ii)}=\sum^N_{n=1}\|P^n_{avg}-\frac{1}{N}\|
\end{aligned}
\end{equation}
where $P^n_{avg}$ is the averaged probability of all valid samples within the $n$-th bin. The uniformizing loss is optimized based only on valid samples, which are not filtered out by auto-masking $\mathcal{M}(\cdot)$. We adopt the uniformizing loss in the following form: 
\begin{equation}
    L_u=\mathcal{M}(L_u^{ii)})
\end{equation}
The final loss $L_{final}$ is computed as:
\begin{equation}
    L_{final}=L_p+\alpha_2 L_{smooth}+\alpha_3 L_u
\end{equation}
where $\alpha_2$ and $\alpha_3$ are weighted factors. 

\begin{wrapfigure}{m}{0.5\textwidth}
% \vspace{-3mm}
\includegraphics[width=0.5\textwidth]{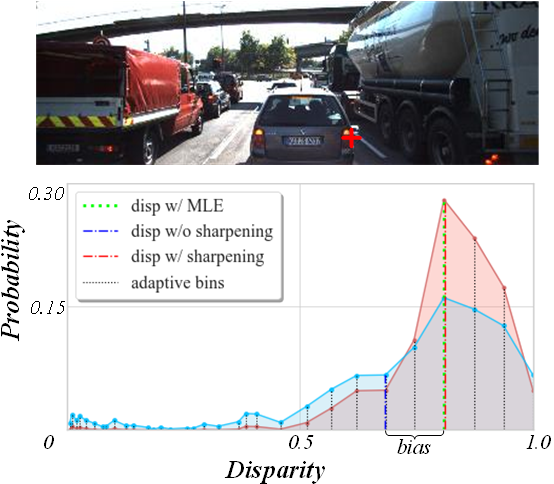}
\caption{\textbf{Benefit of sharpening.} Encouraging distribution to exhibit extremes reduces the bias between \textit{soft-argmax} and MLE.} 
\vspace{-3em}
\end{wrapfigure}

\textbf{Sharpening} reduces the sensitivity of \textit{soft-argmax} to the shape of the distribution by encouraging the presence of a dominant peak in pixel-wise probability distributions. Figure 3 illustrates two probability distributions of the selected pixel: the blue one represents the unsharpened version, while the red is rectified by sharpening. This is achieved by introducing a temperature parameter $\tau\in (0,1]$ into \textit{softmax}, modifying the probability volume as follow: 

\begin{equation}
    P^n_{(u,v)}=\frac{\exp (y^n_{(u,v)}/\tau)}{\sum_{i=1}^N\exp (y^i_{(u,v)}/\tau)}
\end{equation}

\section{Experiments}

\subsection{Implementation Details}
We train the model for $20$ epochs using the Adam optimizer \cite{kingma2014adam} with an initial learning rate of $1e^{-4}$, which decays to $1e^{-5}$ after $15$ epochs. ResNet-18 \cite{he2016deep} is employed as the backbone with weights pretrained on ImageNet \cite{russakovsky2015imagenet}. Experiments are conducted with an image resolution of $416\times 128$ and $jpeg$ format to save memory and computational resources. Data augmentation are applied, including horizontal flipping and color jittering in brightness ($\pm 0.2$), contrast ($\pm 0.2$), saturation ($\pm 0.2$) and hue ($\pm 0.1$). We adopt the hyper-parameters $\alpha_1=0.85$ and $\alpha_2=10^{-3}$, following Monodepth2 \cite{godard2019digging}. The uniformizing term is weighted by $\alpha_3=1$, and the temperature parameter for sharpening is set to $\tau=0.5$.

\subsection{KITTI Results}
We employ the KITTI dataset \cite{geiger2012we}, following the data split of Eigen et al. \cite{eigen2015predicting}, which is commonly used in depth estimation and other computer vision tasks. After pre-processing by \cite{zhou2017unsupervised} to remove static frames, we obtain $39,810$ monocular training triplets, with $4,424$ reserved for validation. During inference, depth estimation is limited to a range of 80m \cite{eigen2015predicting,godard2017unsupervised}, and we address scale ambiguity using pre-image median ground truth scaling \cite{zhou2017unsupervised}. Evaluation of depth estimation is conducted using standard metrics outlined in \cite{eigen2015predicting}.

\begin{table}[htbp]
\footnotesize
    \setlength{\tabcolsep}{0.4mm}{
    \begin{tabular}{|l|c|cccc|ccc|}
    \hline
    & &\multicolumn{4}{c|}{Error metric $\downarrow$} &\multicolumn{3}{c|}{Accuracy metric $\uparrow$}\\
    \cline{3-9}
     &$\#bins$& $Abs rel$ & $Sq Rel$ & $RMSE$  & $RMSE log$ &$\delta < 1.25$ &$\delta < 1.25^2$ &$\delta < 1.25^3$\\             
    \hline
    Monodepth2 \cite{godard2019digging}&-& 0.128& 0.961& 5.137&0.205&0.846&0.951&0.979\\
    + UD&32& 0.125&0.920&4.929& 0.201&0.855& 0.954&0.980\\
    + SID&32& 0.125& 0.970& 5.032&0.200&  0.857& 0.954& 0.980\\
     + ADDV \textbf{(Ours)}&32& \textbf{0.119}& \textbf{0.904}  & \textbf{4.922}  & \textbf{0.197}&  \textbf{0.864}& \textbf{0.956}& 0.980\\
     \hline
    + UD&128& 0.127& 1.011& 5.112& 0.203&  0.852& 0.953& 0.979\\
    + SID&128& 0.123& 0.939  & 4.986  & 0.199&0.858& 0.955& 0.980\\
     + ADDV \textbf{(Ours)}&128& \textbf{0.119}& \textbf{0.919}  & \textbf{4.892}  & \textbf{0.196}&  \textbf{0.866}& \textbf{0.956}& 0.980\\
   
    \hline
     
    \end{tabular}
    }
    \caption{\textbf{Quantitative results.} Comparison of ADDV to existing discretization strategies (UD and SID) on the KITTI benchmark. For error metrics, lower is better, and for accuracy metrics, higher is better. The best results are in \textbf{bold}.}
\end{table}

\textbf{Quantitative} results are recorded in Table 1, with Monodepth2 serving as the baseline. Comparing ADDV to other discretization strategies on KITTI dataset, we observe that our adaptive method is superior to UD or SID across most metrics for the same number of bins. As reported in \cite{fu2018deep}, a higher number of bins enhances the performance of SID, while it also increases network complexity and computational costs. Nevertheless, despite employing only 32 bins, ADDV surpasses the 128-bin SID, underscoring the efficacy of our adaptive approach over handcrafted designs. 

\begin{figure}[!ht]
    \includegraphics[width=\textwidth]{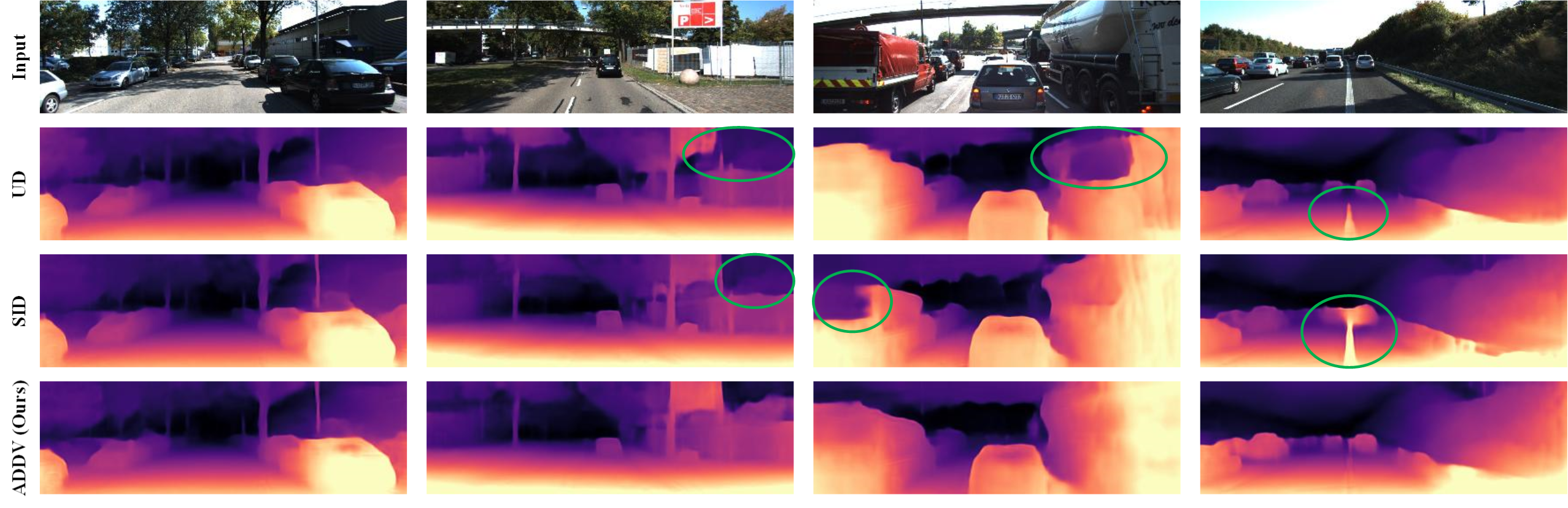}
    \begin{tabular}{@{\hskip 17mm}c@{\hskip 27mm}c@{\hskip 27mm}c@{\hskip 27mm}c}
    \small{a)} & \small{b)} & \small{c)}& \small{d)} \\
    \end{tabular}
    \caption{\textbf{Qualitative results.} All three discretization strategies adopt 32 bins and are evaluated on the validation dataset. Failure cases are highlighted. }
\label{fig4}
\end{figure}

\begin{figure}[t]
    \includegraphics[width=1\textwidth]{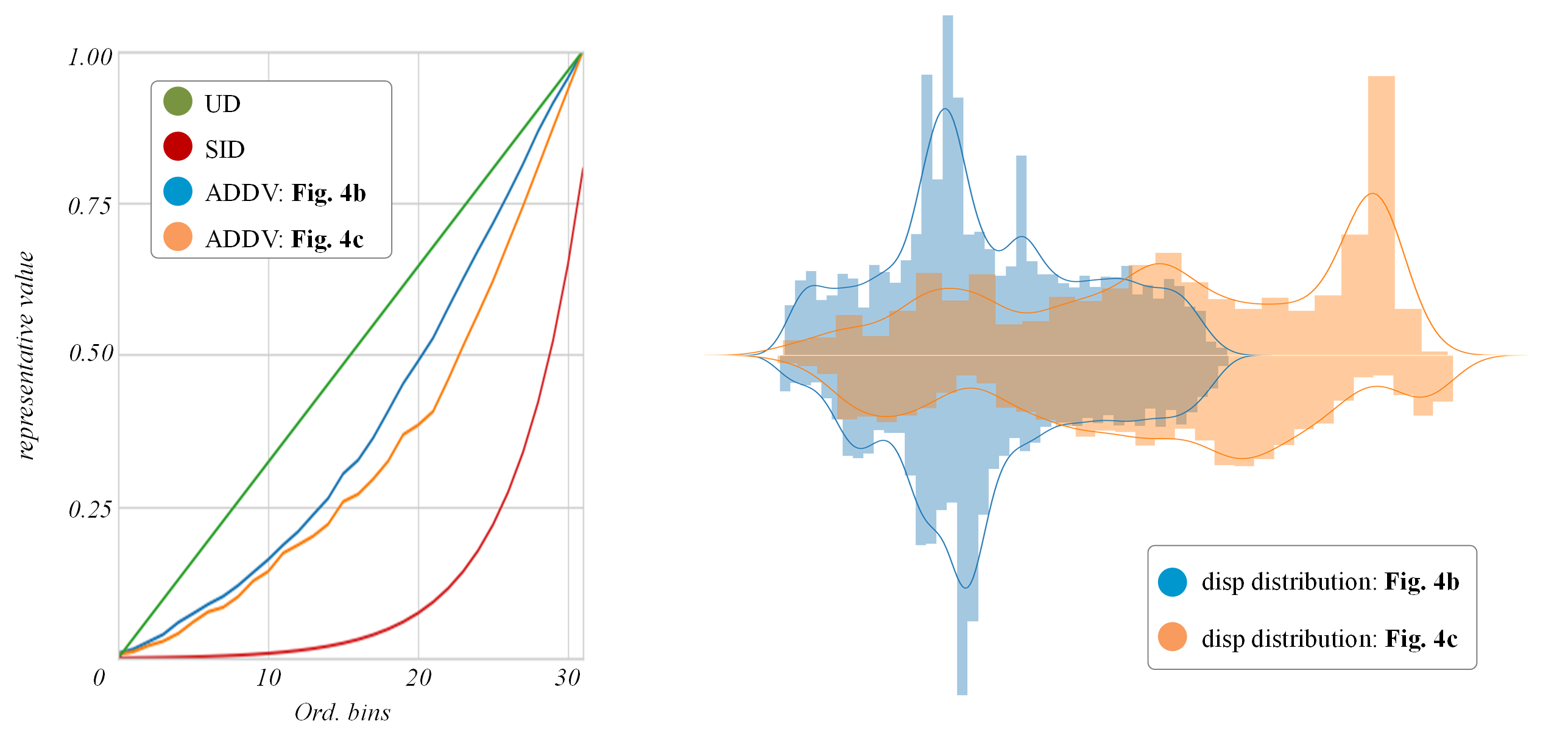} 
    \begin{tabular}{@{\hskip 25mm}c@{\hskip 60mm}c}
    \small{a)} & \small{b)} \\
    \end{tabular}
    \caption{\textbf{Analysis of adaptive bins.} a) Curves of representative values generated by UD (green), SID (red) and ADDV (blue \& orange) for two scenes: Fig 4b and Fig. 4c. b) Histogram of disparity for the same two scenes.}
\label{fig5}
\end{figure}

\textbf{Qualitative} performance is illustrated on four distinct scenes in Figure 4. The model equipped with ADDV produces higher quality depth maps with fewer artifacts and mistakes. Notably, it exhibits robustness in handling confusing patterns such as glassy reflective surfaces (Fig. 4c) and traffic lanes (Fig. 4d). We further analysis the effectiveness of ADDV. The curves in Figure 5a present the representative values generated for Fig. 4b and Fig. 4c. While ADDV dynamically adjusts to fit different scenes, UD and SID generate fixed bins. Furthermore, the curves of ADDV show variability based on input characteristics, aligning with UD for distant scenes (Fig. 4b) and SID for close-up shots (Fig. 4c). This observation emphasizes that a fixed strategy is essentially a specific case of the adaptive method. To illustrate how ADDV learns the depth distribution for various scenes, we display the histogram of ground-truth disparity values and estimation results from ADDV in Figure 5b. The upper part corresponds to the ground truth, while the lower part is from ADDV results. It adeptly captures the depth clues in scenes and mimics depth distributions without any supervision. 

\subsection{Ablation Study}

\begin{wraptable}{r}{0.5\textwidth}
\vspace{-12mm}
    \scriptsize
    \setlength{\tabcolsep}{0.28mm}{
    \begin{tabular}{lcccccc}
    \toprule
    &\textbf{u}  &\textbf{s}     &$\#bins$  & $Abs rel\downarrow$  & $RMSE\downarrow$    &$\delta < 1.25\uparrow$ \\
    \hline
    baseline      &- &-           &-     & 0.128& 5.137&  0.846\\
+ADDV&$\checkmark$ & $\checkmark$& 16    & 0.121& 4.942&  0.858\\ 
+ADDV&$\checkmark$ & $\checkmark$& 32& \textbf{0.119}& 4.922&0.864\\
+ADDV&$\checkmark$ & $\checkmark$& 64& 0.121& \textbf{4.840}&0.860\\
+ADDV&$\checkmark$ & $\checkmark$& 128&\textbf{0.119}& 4.892&\textbf{0.866}\\
\hline
+ADDV&-& -& 32& 0.124& 5.002&0.860\\
+ADDV&$\checkmark$ & -    & 32    & 0.120& 4.925&  0.863\\
+ADDV&-& $\checkmark$   & 32    & -& -&  -\\
   
    \bottomrule
    \end{tabular}}
    \caption{\textbf{Ablation Study.} Quantitative results for different variants of ADDV. \textbf{u}-uniformizing; \textbf{s}-sharpening.}
    \label{tab2}
    \vspace{-5mm}
\end{wraptable}

We explore the influences of uniformizing, sharpening and bins number in this ablation study. 

\textbf{Uniformizing and sharpening} are regularizations employed during network training under self-supervised conditions, thereby preventing performance degradation. The quantitative results in the last few rows of Table 2 illustrate the impact of these strategies. Without uniformizing and sharpening, although the model achieves notable improvements over the baseline, it fails to attain the optimal level. While sharpening can enhance results, it poses a risk of training instability and model collapse. On the other hand, the constraints imposed by uniformizing significantly enhance model performance and contribute to stable training and convergence.

\textbf{Bins number} greatly affects the performance of discretization strategies. We report this influence across four levels in Table 2. ADDV achieves optimal performance with 32 and 128 bins. However, increasing the number of bins also escalates model complexity. Therefore, we recommend maintaining a discretization level of 32 for practical implementation.

\section{Conclusion}
We have presented a novel learnable module, Adaptive Discrete Disparity Volume (ADDV), designed for self-supervised monocular depth estimation within a CNN architecture. This module addresses the limitations of conventional discretization strategies by actively generating customized depth bins tailored to different input scenes, thus eliminating the need for handcrafted design or additional supervision. To cope with the issue of lacking fine-grained constraints during training, we have introduced uniformizing and sharpening. Empirical results demonstrate that even in the absence of explicit supervision, adaptive methods are capable of scene perception and effectively processing global information. It consequently outperforms other strategies, yielding higher quality depth maps. Furthermore, our ablation study confirms the efficacy of both training strategies in improving performance.

Our future work will explore removing the limitation on the number of bins, thereby enhancing the flexibility of the network and enabling it to handle a wider range of scenes. We also plan to leverage uncertainty maps generated by discretization methods to refine depth maps.

\bibliographystyle{splncs04}
\bibliography{ref}



% LNCS Homepage, \url{http://www.springer.com/lncs}. Last accessed 4
% \end{thebibliography}
\end{document}